\documentclass{article}
\usepackage{spconf,amsmath,graphicx}
\usepackage{enumitem}
\setlist{nosep, leftmargin=14pt}
\usepackage{booktabs}

\title{ARIA: On the Interaction between Architectures, Initialization and Aggregation Methods for Federated Visual Classification}
\name{Vasilis Siomos$^1$, Sergio Naval-Marimont$^1$, Jonathan Passerat-Palmbach$^{1,2}$, Giacomo Tarroni$^{1,2}$}
\address{
  $^1$ CitAI Research Centre, Department of Computer Science City, University of London\\
  $^2$ BioMedIA, Department of Computing, Imperial College London
}
\begin{document}
\ninept
\maketitle
\begin{abstract}
Federated Learning (FL) is a collaborative training paradigm that allows for privacy-preserving learning of cross-institutional models by eliminating the exchange of sensitive data and instead relying on the exchange of model parameters between the clients and a server. Despite individual studies on how client models are aggregated, and, more recently, on the benefits of ImageNet pre-training, there is a lack of understanding of the effect the architecture chosen for the federation has, and of how the aforementioned elements interconnect. To this end, we conduct the first joint ARchitecture-Initialization-Aggregation study and benchmark ARIAs across a range of medical image classification tasks. We find that, contrary to current practices, ARIA elements have to be chosen together to achieve the best possible performance. Our results also shed light on good choices for each element depending on the task, the effect of normalization layers, and the utility of SSL pre-training, pointing to potential directions for designing FL-specific architectures and training pipelines.
\end{abstract}
\begin{keywords}
Federated Learning, Self-Supervised Pre-training
\end{keywords}
%

\begin{table*}[t]
\centering
\caption{Average balanced accuracy across 6 clients on Fed-ISIC. IN top-1 accuracy reported next to model name. Models listed in decreasing measured training throughput (using AMP). Difference from average balanced accuracy of centrally trained model in parentheses.}
\label{tab:fedisic}
\resizebox{\textwidth}{!}{%
\begin{tabular}{@{}l|lll|lll|lll@{}}
\toprule
Initialization           & \multicolumn{3}{c|}{Random}                                                               & \multicolumn{3}{c|}{ImageNet Pre-Training}                                              & \multicolumn{3}{c}{DINO on Skin SSL dataset}                                              \\ \midrule
Agg. Method              & FedAvg                    & FedOpt                    & SCAFFOLD                          & FedAvg                            & FedOpt                   & SCAFFOLD                 & FedAvg                    & FedOpt                    & SCAFFOLD                          \\ \midrule
ResNet-18 (69.76)        & 51.65 ($\downarrow$ 9.8)  & 46.7 ($\downarrow$ 14.7)  & 52.45 ($\downarrow$ 9)            & 65.87 ($\downarrow$ 4.3)          & 67.55 ($\downarrow$ 2.6) & 68.66 ($\downarrow$ 1.5) & 66.57 ($\downarrow$ 5.7)  & 62.36 ($\downarrow$ 10)   & 66.87 ($\downarrow$ 5.4)          \\
NF-ResNet-50 (80.64)     & 55.93 ($\downarrow$ 6.1)  & 56.25 ($\downarrow$ 5.8)  & \textbf{59.64 ($\downarrow$ 2.4)} & 71.88 ($\uparrow$ 0.9)            & 68.75 ($\downarrow$ 2.2) & 71.53 ($\uparrow$ 0.5)   & 67.83 ($\downarrow$ 0.7)  & 67.92 ($\downarrow$ 0.6)  & 70.11 ($\uparrow$ 1.6)            \\
ResNet-50 (80.86)        & 49.11 ($\downarrow$ 12)   & 46.91 ($\downarrow$ 14.2) & 48.13 ($\downarrow$ 13)           & 67.97 ($\downarrow$ 6.3)          & 66.16 ($\downarrow$ 8.1) & 68.48 ($\downarrow$ 5.8) & 65.16 ($\downarrow$ 7.2)  & 66.46 ($\downarrow$ 5.9)  & 66.34 ($\downarrow$ 6)            \\
WRN-50-2 (81.6)          & 50.53 ($\downarrow$ 8)    & 50.12 ($\downarrow$ 8.4)  & 51.03 ($\downarrow$ 7.5)          & 69.54 ($\downarrow$ 5.3)          & 67.68 ($\downarrow$ 7.2) & 70.34 ($\downarrow$ 4.5) & 65.56 ($\downarrow$ 6.9)  & 64.22 ($\downarrow$ 8.3)  & 66.66 ($\downarrow$ 5.8)          \\
DenseNet-121 (74.43)     & 49.42 ($\downarrow$ 13.3) & 45.95 ($\downarrow$ 16.8) & 52.79 ($\downarrow$ 9.9)          & 67.34 ($\downarrow$ 5.8)          & 68.03 ($\downarrow$ 5)   & 68.52 ($\downarrow$ 4.6) & 66.28 ($\downarrow$ 5.3)  & 64.94 ($\downarrow$ 6.6)  & 67.38 ($\downarrow$ 4.2)          \\
SWIN-T (81.47)           & 45.73 ($\uparrow$ 23.2)   & 44.13 ($\uparrow$ 21.6)   & 45.00 ($\uparrow$ 22.5)           & 71.19 ($\downarrow$ 1.3)          & 71.81 ($\downarrow$ 0.6) & 73.13 ($\uparrow$ 0.7)   & 72.13 ($\uparrow$ 1.7)    & 71.40 ($\uparrow$ 0.9)    & 73.77 ($\uparrow$ 3.3)            \\
EfficientNetV2-S (84.22) & 46.59 ($\downarrow$ 10.8) & 46.59 ($\downarrow$ 10.8) & 47.51 ($\downarrow$ 9.8)          & 70.00 ($\downarrow$ 9.6)          & 71.48 ($\downarrow$ 8.1) & 73.18 ($\downarrow$ 6.4) & 57.99 ($\downarrow$ 14.9) & 59.74 ($\downarrow$ 13.1) & 64.98 ($\downarrow$ 7.9)          \\
ViT-B-16 (81.07)         & 47.84 ($\uparrow$ 7.2)    & 49.52 ($\uparrow$ 8.9)    & 48.44 ($\uparrow$ 7.8)            & 65.86 ($\uparrow$ 1.6)            & 65.18 ($\uparrow$ 0.9)   & 68.09 ($\downarrow$ 3.8) & 71.06 ($\downarrow$ 2.9)  & 71.52 ($\downarrow$ 2.5)  & 69.49 ($\downarrow$ 4.5)          \\
ConvNext-S (83.61)       & 48.10 ($\downarrow$ 7.9)  & 49.93 ($\downarrow$ 6.1)  & 48.56 ($\downarrow$ 7.5)          & \textbf{75.08 ($\downarrow$ 0.1)} & 73.40 ($\downarrow$ 1.7) & 74.28 ($\downarrow$ 0.8) & 72.07 ($\downarrow$ 3)    & 73.57 ($\downarrow$ 1.5)  & \textbf{74.56 ($\downarrow$ 0.5)} \\ \bottomrule
\end{tabular}%
}
\end{table*}

\begin{table*}[t]
\centering
\caption{Average accuracy across 4 clients on OrganAMNIST with $\alpha=0.1$. IN top-1 accuracy reported next to model name. Models listed in decreasing measured training throughput (using AMP). Difference from the accuracy of the centrally trained model in parentheses.}
\label{tab:01mnist}
\resizebox{\textwidth}{!}{%
\begin{tabular}{@{}l|lll|lll|lll@{}}
\toprule
Initialization           & \multicolumn{3}{c|}{Random}                                                            & \multicolumn{3}{c|}{ImageNet Pre-Training}                                           & \multicolumn{3}{c}{DINO on Abdomen-SSL}                                                \\ \midrule
Agg. Method              & FedAvg                   & FedOpt                           & SCAFFOLD                 & FedAvg                  & FedOpt                  & SCAFFOLD                         & FedAvg                   & FedOpt                           & SCAFFOLD                 \\ \midrule
ResNet-18 (69.76)        & 88.8 ($\downarrow$5.6)   & 90.76 ($\downarrow$3.6)          & 89.16 ($\downarrow$5.2)  & 94.02 ($\downarrow$1.9) & 94.78 ($\downarrow$1.2) & 94.33 ($\downarrow$1.6)          & 83.54 ($\downarrow$9.8)  & 87.89 ($\downarrow$5.5)          & 84.76 ($\downarrow$8.6)  \\
NF-ResNet-50 (80.64)     & 71.6 ($\downarrow$16.3)  & 78.84 ($\downarrow$9.1)          & 73.8 ($\downarrow$14.1)  & 94.39 ($\downarrow$1.4) & 95.26 ($\downarrow$0.5) & 95.2 ($\downarrow$0.6)           & 84.58 ($\downarrow$7.9)  & 87.93 ($\downarrow$4.5)          & 86.92 ($\downarrow$5.5)  \\
ResNet-50 (80.86)        & 83.32 ($\downarrow$10.5) & 86.6 ($\downarrow$7.2)           & 84.82 ($\downarrow$9.0)  & 91.98 ($\downarrow$3.5) & 92.98 ($\downarrow$2.5) & 92.32 ($\downarrow$3.1)          & 81.33 ($\downarrow$12.9) & 85.69 ($\downarrow$8.5)          & 81.49 ($\downarrow$12.8) \\
WRN-50-2 (81.6)          & 84.52 ($\downarrow$9.6)  & 85.58 ($\downarrow$8.5)          & 83.82 ($\downarrow$10.3) & 90.56 ($\downarrow$4.3) & 91.71 ($\downarrow$3.2) & 90.4 ($\downarrow$4.5)           & 79.98 ($\downarrow$13.7) & 85.02 ($\downarrow$8.6)          & 77.09 ($\downarrow$16.5) \\
DenseNet-121 (74.43)     & 86.01 ($\downarrow$8.6)  & 89.12 ($\downarrow$5.5)          & 85.06 ($\downarrow$9.6)  & 94.72 ($\downarrow$2.2) & 95.1 ($\downarrow$1.9)  & 94.68 ($\downarrow$2.3)          & 85.26 ($\downarrow$9.2)  & 89.21 ($\downarrow$5.3)          & 84.94 ($\downarrow$9.5)  \\
SWIN-T (81.474)          & 83.03 ($\downarrow$8.6)  & 85.17 ($\downarrow$6.4)          & 83.16 ($\downarrow$8.4)  & 95.64 ($\downarrow$0.6) & 95.83 ($\downarrow$0.4) & 95.83 ($\downarrow$0.4)          & 83.4 ($\downarrow$8.2)   & 86.4 ($\downarrow$5.2)           & 84.8 ($\downarrow$6.8)   \\
EfficientNetV2-S (84.22) & 88.8 ($\downarrow$6.2)   & \textbf{91.46 ($\downarrow$3.6)} & 89.19 ($\downarrow$5.9)  & 94.0 ($\downarrow$2.7)  & 94.26 ($\downarrow$2.4) & 93.46 ($\downarrow$3.2)          & 61.19 ($\downarrow$31.6) & 67.54 ($\downarrow$25.3)         & 56.2 ($\downarrow$36.6)  \\
ViT-B-16 (81.072)        & 83.14 ($\downarrow$4.2)  & 83.52 ($\downarrow$3.9)          & 83.85 ($\downarrow$3.5)  & 95.3 ($\downarrow$1.5)  & 95.96 ($\downarrow$0.9) & \textbf{96.01 ($\downarrow$0.8)} & 81.34 ($\downarrow$6.8)  & 83.76 ($\downarrow$4.4)          & 81.99 ($\downarrow$6.2)  \\
ConvNext-S (83.61)       & 53.76 ($\downarrow$35.4) & 56.07 ($\downarrow$33.1)         & 55.34 ($\downarrow$33.8) & 94.12 ($\downarrow$2.6) & 94.92 ($\downarrow$1.8) & 94.84 ($\downarrow$1.9)          & 87.31 ($\downarrow$6.0)  & \textbf{89.68 ($\downarrow$3.7)} & 87.64 ($\downarrow$5.7)  \\ \bottomrule
\end{tabular}%
}
\end{table*}

\section{Introduction}
\label{sec:intro}

Federated learning (FL) for healthcare \cite{sheller2020federated} has emerged as a promising approach that enables collaborative machine learning without direct access to raw patient data. The typical scenario for medical imaging data is the cross-silo setting, where a small number of data owners/stakeholders fully participate in a round of federated training by training their local/client models and sending the parameters to a central server, which aggregates the client models into a server/global model. The global model is then broadcast to all clients to start the next round, until training stops, and the final model is delivered to the stakeholders for deployment.

Since the seminal FedAvg paper \cite{fedavg}, progress in cross-silo visual classification has been hard to determine, with innovation often focusing on improving the aggregation strategy for the frequent scenario where the client datasets are heterogeneous \cite{hsu2019measuring,scaffold}. Unfortunately, proposed methods most commonly use randomly initialized model weights, small/toy models, or both \cite{pieri2023handling}. This makes comparing and drawing conclusions for real-world medical settings difficult.

Recent studies \cite{chen2022importance, nguyen2022begin} have been exploring the value of using ImageNet (IN) pre-trained networks for FL, showcasing improvements in closing the gap to centralized performance, improving overall performance, and reducing the effect of data heterogeneity. Another study by Qu et al. \cite{vitfl} highlighted the benefits of using IN pre-trained transformers for FL. Very recently, Pieri et al. \cite{pieri2023handling} focused on the interaction of aggregation methods and architectures, but only examined IN pre-trained weights.

It’s important to note IN pre-training restricts the input to 224x224 RGB images. When up-sampling of the original images is required to achieve that, it leads to a bigger than necessary computational and memory load, and the introduction of aliasing artifacts (e.g. Fig. 1). When down-sampling is required instead, it can degrade performance. Hence, IN pre-training is not a silver bullet, and benchmarking architectures and aggregation strategies without pre-training is also important. Furthermore, task-relevant pre-training through self-supervised learning (SSL) has recently emerged as a highly-effective alternative to IN pre-training \cite{bob}, but its usefulness in the FL setting remains largely unexplored.

Motivated by the above, we conduct what, to the best of our knowledge, is the first study to jointly examine ARIAs: Architecture-Initialization-Aggregation combinations: we select 9 architectures, with the weights initialized from 3 starting points (Random, ImageNet, SSL on a relevant dataset), and use 3 of the most common methods (FedAvg, FedOpt, SCAFFOLD) to aggregate the models. We focus on perhaps the most beneficial domain for FL, medical imaging, and evaluate the resulting ARIAs on 3 different medical imaging datasets, namely Fed-ISIC, and two versions of OrganAMNIST (with and without simulated heterogeneity).

Our results after training more than 300 ARIAs indicate to researchers and practitioners designing FL pipelines for medical imaging data that all elements of an ARIA have to be evaluated together, but also shed light on the individual effects of network size, normalization methods, architecture choice, and utility of SSL pre-training.

\section{Methods}
\label{sec:methods}

\subsection{(AR)chitectures}
We aim to compare popular architectures from both the convolutional and transformer families, while also pinpointing architectural block that boost FL performance. All models are of reasonable size and throughput for our target tasks. For comparison, rows in tables \ref{tab:fedisic},\ref{tab:01mnist},\ref{tab:100mnist} are listed in decreasing training throughput, and we include model parameter counts here. From the family of residual networks \cite{resnet}, we choose a ResNet-18 (11.7M parameters), a ResNet-50 (25.6M), and a Wide-ResNet-50-2 \cite{wrn} (68.9M), to examine the effect of depth and width. A DenseNet-121 \cite{densenet} (8M), shows how the density of residual connections and feature re-use affect performance. 

These architectures employ Batch Normalisation (BN), which is known to degrade FL performance in heterogeneous settings, due to the BN statistics being averaged across heterogeneous image distributions\cite{makingbn}. There is no clear solution to this, with replacing BN layers in a ResNet with Group or Layer Normalization, or not sharing the BN layers, having been proposed before \cite{makingbn}. Our approach to providing insight into alternatives to BN is benchmarking networks that altogether do not use BN in their original form. To this end, we use a Normalization-Free (NF) ResNet-50 \cite{nfresnet} (25.6M). NF architectures rely on Scaled Weight Standardization (SWS), i.e. careful scaling of weights, instead of normalization, to achieve correct signal propagation during learning.

Since the emergence of vision transformers, more advanced convolutional architectures have been introduced with the goal of outperforming them, and we pick EfficientNetV2-S \cite{efficientnetv2} (21.5M, uses BN), an evolution of ResNets guided by neural architecture search, and ConvNext-S \cite{convnext} (50.2M, uses LN), which borrows design principles from SWIN transformers, as modern CNN benchmarks.

From the transformer family, we benchmark a ViT-B-16 \cite{vit} (86.6M) and a SWIN-T \cite{swin} (28.3M), to compare convolution with self-attention. Both employ Layer Normalization (LN).

\subsection{(I)initialization}
\begin{figure}[tb]
\centering
\includegraphics[width=0.9\columnwidth]{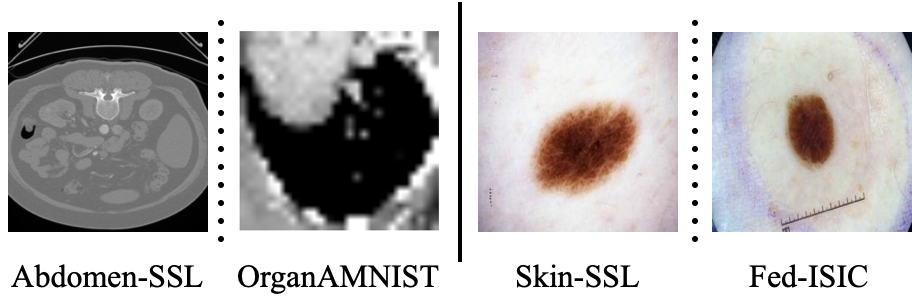}
\hfill
\caption{Samples from the SSL and respective target datasets.}
\label{fig:ssl_datasets}
\end{figure}

Both random and IN initializations are of interest depending on the application, as explained in section 1; all rows in rows in tables \ref{tab:fedisic},\ref{tab:01mnist},\ref{tab:100mnist} list the pre-trained model's top-1 accuracy on IN. However, in medical imaging scenarios, it is often the case that i) the target task images are dissimilar to the natural ones of IN and ii) medical datasets with similar images are publicly available. This leads us to examine whether training the models using self-supervised learning (SSL) as a pre-cursor task can be a beneficial initialization strategy for FL. We construct two task-relevant pre-training datasets, Abdomen-SSL and Skin-SSL (Fig \ref{fig:ssl_datasets}), and train all models using DINO \cite{DINO}for 100/300 epochs on the two datasets respectively, with the length chosen based on the loss plateauing.

\subsection{(A)ggregation methods}
We limit our scope to methods that produce a global model $w_g$, and not a personalized model for each client. We select three of the most common aggregation strategies, namely FedAvg, FedOpt, and SCAFFOLD, which share in common their ease/lack of hyper-parameters to be tuned, allowing for more universal insights. \textbf{FedAvg} \cite{fedavg} is the seminal FL parameter averaging method, which uses as the sample-weighted average of client models $w_i$ to produce $w_g=N_i/N\cdot\sum_{i=1}^Cw_i$. 

The \textbf{FedOpt} \cite{fedopt} family of methods de-couples server and client-side optimization, and the server can employ any optimizer like SGD, Adam, etc. We use SGD with momentum at the server, similar to FedAvgM \cite{hsu2019measuring}, with the addition of a cosine annealing to the server learning rate. The server learning rate is 1.0, tuned from \{0.5, 1\}, and the momentum to 0.6, tuned from \{0.6, 0.9\}. 

\textbf{SCAFFOLD} \cite{scaffold} utilizes control variates to correct local model updates against client-drift (divergence from the global model). These parameters are of equal size to the model and there is one set being stored locally and another exchanged alongside the model, leading to twice the communication and triple the local storage cost.

\section{Experiments}
\label{sec:experiments}

\begin{table*}[t]
\centering
\caption{Average accuracy across 4 clients on OrganAMNIST with $\alpha=100$. IN top-1 accuracy reported next to model name. Models listed in decreasing measured training throughput (using AMP). Difference from accuracy of centrally trained model in parentheses.}
\label{tab:100mnist}
\resizebox{\textwidth}{!}{%
\begin{tabular}{@{}l|lll|lll|lll@{}}
\toprule
Initialization           & \multicolumn{3}{c|}{Random}                                                         & \multicolumn{3}{c|}{ImageNet Pre-Training}                                          & \multicolumn{3}{c}{DINO on Abdomen-SSL}                                              \\ \midrule
Agg. Method              & FedAvg                  & FedOpt                         & SCAFFOLD                 & FedAvg                  & FedOpt                          & SCAFFOLD                & FedAvg                  & FedOpt                           & SCAFFOLD                \\ \midrule
ResNet-18 (69.76)        & 93.8 ($\downarrow$0.6)  & 94.3 ($\downarrow$0.1)         & 93.97 ($\downarrow$0.4)  & 96.05 ($\uparrow$0.1)   & 96.38 ($\uparrow$0.4)           & 95.99 ($\downarrow$0.0) & 92.06 ($\downarrow$1.3) & 93.47 ($\uparrow$0.1)            & 92.14 ($\downarrow$1.2) \\
NF-ResNet-50 (80.64)     & 84.28 ($\downarrow$3.6) & 88.09 ($\uparrow$0.2)          & 84.4 ($\downarrow$3.5)   & 95.5 ($\downarrow$0.3)  & 95.64 ($\downarrow$0.1)         & 95.6 ($\downarrow$0.2)  & 92.08 ($\downarrow$0.4) & 92.74 ($\uparrow$0.3)            & 92.09 ($\downarrow$0.4) \\
ResNet-50 (80.86)        & 93.39 ($\downarrow$0.5) & 94.0 ($\uparrow$0.2)           & 93.54 ($\downarrow$0.3)  & 94.98 ($\downarrow$0.5) & 95.56 ($\uparrow$0.1)           & 95.34 ($\downarrow$0.1) & 92.8 ($\downarrow$1.4)  & \textbf{93.51 ($\downarrow$0.7)} & 92.69 ($\downarrow$1.5) \\
WRN-50-2 (81.6)          & 93.74 ($\downarrow$0.3) & 93.99 ($\downarrow$0.1)        & 93.72 ($\downarrow$0.4)  & 94.7 ($\downarrow$0.2)  & 95.42 ($\uparrow$0.5)           & 94.76 ($\downarrow$0.1) & 92.24 ($\downarrow$1.4) & 93.1 ($\downarrow$0.5)           & 92.52 ($\downarrow$1.1) \\
DenseNet-121 (74.43)     & 93.95 ($\downarrow$0.7) & 94.28 ($\downarrow$0.3)        & 93.66 ($\downarrow$1.0)  & 96.53 ($\downarrow$0.4) & \textbf{97.0 ($\downarrow$0.0)} & 96.66 ($\downarrow$0.3) & 93.4 ($\downarrow$1.1)  & 94.11 ($\downarrow$0.4)          & 93.38 ($\downarrow$1.1) \\
SWIN-T (81.47)           & 90.64 ($\downarrow$1.0) & 90.89 ($\downarrow$0.7)        & 90.27 ($\downarrow$1.3)  & 96.61 ($\uparrow$0.4)   & 96.82 ($\uparrow$0.6)           & 96.6 ($\uparrow$0.4)    & 89.66 ($\downarrow$2.0) & 90.86 ($\downarrow$0.8)          & 89.68 ($\downarrow$1.9) \\
EfficientNetV2-S (84.22) & 94.84 ($\downarrow$0.2) & \textbf{95.13 ($\uparrow$0.1)} & 94.96 ($\downarrow$0.1)  & 96.22 ($\downarrow$0.5) & 96.48 ($\downarrow$0.2)         & 96.28 ($\downarrow$0.4) & 89.18 ($\downarrow$3.6) & 92.03 ($\downarrow$0.8)          & 89.02 ($\downarrow$3.8) \\
ViT-B-16 (81.07)         & 86.42 ($\downarrow$1.0) & 86.34 ($\downarrow$1.0)        & 86.54 ($\downarrow$0.8)  & 96.12 ($\downarrow$0.7) & 96.3 ($\downarrow$0.5)          & 96.25 ($\downarrow$0.6) & 86.67 ($\downarrow$1.5) & 87.61 ($\downarrow$0.6)          & 86.94 ($\downarrow$1.2) \\
ConvNext-S (83.61)       & 84.56 ($\downarrow$4.6) & 87.29 ($\downarrow$1.9)        & 78.48 ($\downarrow$10.7) & 96.3 ($\downarrow$0.4)  & 96.24 ($\downarrow$0.5)         & 96.18 ($\downarrow$0.5) & 92.15 ($\downarrow$1.2) & 92.87 ($\downarrow$0.5)          & 92.24 ($\downarrow$1.1) \\ \bottomrule
\end{tabular}%
}
\end{table*}

\subsection{Datasets}
We conduct experiments on abdominal CT with OrganAMNIST \cite{medmnist} and skin lesions with Fed-ISIC \cite{flamby}. The latter is naturally federated with multi-center data, and for the former we construct a federated version of 4 clients, by following convention and using the Dirichlet partitioning strategy \cite{hsu2019measuring}, which induces size and label distribution heterogeneity based on the controllable concentration parameter $\alpha$. We examine an IID setting by setting $\alpha=100$, and a highly heterogeneous one by setting $\alpha=0.1$. This leads to a wide range of difficulty to benchmark the chosen models, from the grayscale IID OrganAMNIST to the highly imbalanced, both in label distribution and data size RGB, Fed-ISIC. Moreover, the distance between the domains and ImageNet provides more insight into learning dynamics for the medical community compared to testing on natural images. 

\textbf{OrganAMNIST} \cite{medmnist} consists of 58,850 28x28 grayscale images with 11 organ labels segmented from axial slices of abdominal CT scans. We upscale the images to 224x224 and copy the channel over 3 times for compatibility with IN pre-trained models. Each client has a training and validation set, with the local validation set used to determine good local training hyper-parameters. After that, clients train on the union of their two sets, and accuracy is reported on the original, pooled, test dataset of 17,778 images.

The \textbf{Abdomen-SSL} dataset was created by extracting 20 slices around the center of each volume in 4 abdominal CT datasets \cite{abdomen_ssl_1,abdomen_ssl_2,abdomen_ssl_3}, cropping around the subject, resizing to 224x224 and copying the channel over, resulting in $\sim 21,000$ whole abdomen images. As seen in Fig.1, Abdomen-SSL is quite different to OrganAMNIST. Due to OrganAMNIST’s uniqueness, it is difficult to design a more similar source dataset. However, SSL pre-training can still help the models learn general organ structures and channel redundancy.

\textbf{Fed-ISIC} \cite{flamby} consists of 23,247 RGB skin lesion images with 8 classes, split across 6 clients representing different datacenters and imaging technologies. Fed-ISIC exhibits very high heterogeneity in size and label imbalance, so performance is measured through balanced accuracy, defined as the average recall on each class. We follow the pre-processing in \cite{flamby}, applying color constancy, and centre-cropping while maintaining the aspect ratio. 

\textbf{Skin-SSL} was created from 3 skin lesion datasets \cite{skin_ssl_1,skin_ssl_3,isic2020}, with the largest contributor being ISIC-2020, which has no overlap with Fed-ISIC, and consists predominantly of benign samples.

For all settings, besides federated training, we also train a central model on the pooled datasets to compare the FL models against, and tables \ref{tab:fedisic},\ref{tab:01mnist},\ref{tab:100mnist} present each model's difference from its centrally trained counterpart in parentheses.

\subsection{Hyper-parameters}
In order to concentrate on the ARIA effects, we limit our scope to shared hyper-parameters between the clients (no client-level tuning), and across aggregation methods. For Fed-ISIC, we follow \cite{flamby} and train for 20 rounds, using Weighted Focal Loss, a batch size of 64, Adam with $lr=5\cdot10^{-4}$, and a cosine annealer. Instead of local epochs, each client performs 200 local steps, tuned from [100,200,600], which allows the biggest client to iterate through all of its data, but keeps client drift to a minimum. While 200 steps worked best for SSL and IN pre-training, we use 600 local steps when training from scratch as a \textit{parity measure} since these models have not seen any data prior, and this improved performance. Adam, surprisingly, worked well for all networks, outperforming both SGD with momentum (favors CNNs) and AdamW (favors transformers) in our tests. The learning rate was tuned in the range $[10^{-4},10^-{3}]$; the combination of adaptive momentum buffers at each client and cosine annealing led to different initial learning rates having minimal effect. For OrganAMNIST, we used the local validation sets of the IID partition and majority voting to decide on the use of momentum SGD with ($lr=0.01$, $m=0.9$), a cosine annealing schedule, a batch size of 128, and 50 local steps. We transfer these settings to the heterogeneous case, since the heterogeneity is typically not known about in advance. Results are averaged across two seeds. We open source our code\footnote{https://github.com/siomvas/ARIA}, which uses NVFlare \cite{nvflare}.

\section{Results and Discussion}
\label{sec:results}

\subsection{Comparing initializations}
In the IID OrganAMNIST experiment (Table \ref{tab:100mnist}) IN pretrained networks virtually solve the task, and achieve very low gaps compared to centralized training (max 0.6\%). This gap increases as heterogeneity in the other two datasets increases, as expected, but overall the IN Initialization outperformed the others. This leads to our first important finding: \textbf{ImageNet is generally the best initialization for federated learning on medical datasets}. In SSL initialization, Skin-SSL pre-training is predictably more useful (Table \ref{tab:fedisic}) than Abdomen-SSL due to the source and target images being much more similar (Figure \ref{fig:ssl_datasets}). Abdomen SSL pre-training reduce performance on average, but helped "prime" ConvNext-S and NF-ResNet-50 compared to random initialization, indicating that SSL can counteract the reduction in regularization due to not using BN. Overall, \textbf{Skin-SSL greatly increases the performance of all models compared to random initialization}. Moreover, despite the much shorter pre-training time compared to IN, the SSL initialized ConvNext-S with SCAFFOLD nearly achieves the best overall performance. In summary, our findings suggest SSL can be extremely beneficial in medical FL in multiple scenarios, from task-specific architectures that have no public IN pre-trained weights, to tasks that cannot adhere to IN image size, and potentially even tasks beyond visual classification, such as segmentation. 

\subsection{ResNet depth, width, and connection density}

Despite deepening and widening generally improving the centrally trained model, the increased central training accuracy was not transferred to FL training. Hence, in our findings, \textbf{ResNets do not scale well in FL tasks}, as ResNet-18 outperforms its deeper and wider counterparts in all settings except for Fed-ISIC with IN weights (Table \ref{tab:fedisic}), where the much larger and slower WRN-50-2 is modestly better. If a low memory footprint is a priority, DenseNet-121, which has much fewer parameters than all other networks but lower throughput than other residual networks, performs just as well or better depending on the task, suggesting that its salient characteristic, \textbf{feature re-use, is beneficial for FL.}

\subsection{Comparing normalization methods}

It has been widely discussed, most recently in \cite{makingbn}, that BN impedes FL performance under heterogeneous settings due to the local clients calculating statistics that are not representative of each other’s datasets. Simultaneously, BN is reliant on the batch size being sufficiently big to accurately approximate the mean and variance, in contrast to LN and SWS. For OrganAMNIST we use a batch size of 128; as a result, we observe (Tables \ref{tab:01mnist}, \ref{tab:100mnist}) that \textbf{randomly initialized BN models outperform LN and SWS ones}, under both IID and non-IID distributions, but there is no difference when the models are pre-trained. For Fed-ISIC, where the batch size was 64, BN could not help random models as much, and \textbf{when using IN weights the top three models all use LN or SWS.} 

Compared to the (generally bigger) LN networks, the performance of NF-ResNet (which uses SWS) does not suffer as much for the random initialization, and the model even performs the best out of all random models on Fed-ISIC. \textbf{This makes NF networks and SWS even more promising for FL applications.}

\subsection{Transformers vs CNNs for FL}
\textbf{Randomly initialized transformers perform poorly} in our experiments, a finding that is perhaps due their lack of inductive bias compared to CNNs, and one that we cannot attribute to model size or speed, as SWIN-T often outperformed ViT-B while being similar to ResNet-50 in size. \textbf{Performance between IN initialized transformers and CNNs was very similar}, with the latter being, on average, marginally better when using IN weights. Thus, we find the extra space and time to train a VIT-B-16 model compared to ResNet-50 mostly fruitless. \textbf{SSL initialization greatly increases transformer performance}, and outperforms the IN one in Fed-ISIC. This is despite our SSL pipeline not being tuned to its full extent, which can likely further increase SSL performance. Hence, we argue that transformer models have a place in medical FL applications where the target domain is dissimilar to ImageNet, and suitable datasets to conduct SSL are available.

\subsection{Comparing aggregation methods}
In OrganAMNIST, FedOpt, on average, increases test accuracy by 0.68\% and 2.4\% compared to FedAvg for the IID and non-IID case respectively, while the difference between SCAFFOLD and FedAvg is negligible. In Fed-ISIC, FedOpt led, on average, to a loss of 0.59\% balanced accuracy, but SCAFFOLD consistently improved performance (1.32\% on average), meaning that \textbf{if the extra memory and bandwidth are not an issue, SCAFFOLD is worth considering}. This is in line with its design being for heterogeneous cross-silo, full participation settings, like ours. Despite that, a very important result is that \textbf{the best ARIA uses FedAvg} (Table \ref{tab:fedisic}, IN pre-training). Overall, \textbf{we found the benefits of switching architectures greater than those of switching aggregation methods}, suggesting we need to re-examine how much we have progressed on the algorithmic front since the introduction of FedAvg.

\section{Conclusion}
\label{sec:conclusion}

We conduct the first comprehensive study on ARIAs for federated cross-silo visual classification, giving answers to which parts of an ARIA are most important, and how choices for each compare between them. We find and present evidence that shows FedAvg is still not definitively surpassed, that transformers are not better than CNNs despite recent claims, that IN initialization is beneficial, and, in its absence/non-applicability, SSL also improves performance, as well as the interconnection between these elements. Our work can inform practitioners in the cross-silo setting on which ARIA to employ in real-world scenarios.


\section{Compliance with ethical standards}
\label{sec:ethics}

IEEE-ISBI supports the standard requirements on the use of animal 
This research study was conducted retrospectively using human subject data made available in open access by \cite{medmnist,flamby, abdomen_ssl_1,abdomen_ssl_2,abdomen_ssl_3,skin_ssl_1,skin_ssl_3,isic2020}. Ethical approval was not required as confirmed by the license attached with the open access data.

\section{Acknowledgments}
\label{sec:acknowledgments}

The authors have no relevant interests to disclose.

\bibliographystyle{IEEEbib}
\bibliography{strings,refs}

\end{document}